\documentclass[11pt,a4paper]{article}
\usepackage[hyperref]{acl2020}
\usepackage{times}
\usepackage{latexsym}
\usepackage{booktabs}
\usepackage{algpseudocode}
\usepackage{tikz}
\usepackage{tikz-qtree}
\usepackage{amsfonts}
\usepackage{enumitem}

\usepackage{microtype}
\usepackage{graphicx}
\usepackage{tikz}
\usetikzlibrary{bayesnet}

\aclfinalcopy 

\newcommand{\Cat}{\textsc{Cat}}
\newcommand{\CRF}{\textsc{CRF}}
\newcommand{\entropy}{\mathbb{H}}
\newcommand{\expect}{\mathbb{E}}

\newcommand{\reals}{\mathbb{R}}
\newcommand{\LSE}{\mathrm{LSE}}

\title{Torch-Struct: \\ Deep Structured Prediction Library}

\author{Alexander M. Rush \\
  Cornell University \\
  Department of Computer Science \\
  \texttt{arush@cornell.edu}}

\date{}

\begin{document}
\maketitle

\begin{abstract}

The literature on structured prediction for NLP describes a rich collection of
distributions and algorithms over sequences, segmentations, alignments, and trees; 
however, these algorithms are difficult to utilize in deep learning frameworks.
We introduce Torch-Struct, a library for structured prediction designed to take advantage of and integrate with vectorized, auto-differentiation based frameworks. Torch-Struct includes a broad collection of probabilistic structures accessed through a simple and flexible distribution-based API that connects to any deep learning model. The library utilizes batched, vectorized operations and exploits auto-differentiation to produce readable, fast, and testable code. Internally, we also include a number of general-purpose optimizations to provide cross-algorithm efficiency. Experiments show significant performance gains over fast baselines and 
case-studies demonstrate the benefits of the library. Torch-Struct is available at \url{https://github.com/harvardnlp/pytorch-struct}.


\end{abstract}

\section{Introduction}

Structured prediction is an area of machine learning
focusing on representations of spaces with combinatorial structure, and algorithms for inference and parameter estimation over these structures. Core methods include both tractable exact approaches like dynamic programming and spanning tree algorithms as well as heuristic techniques such linear programming relaxations and greedy search. 

\begin{figure}
    \centering
    \includegraphics[width=0.9\linewidth, trim=1.5cm 1cm 0 0 , clip]{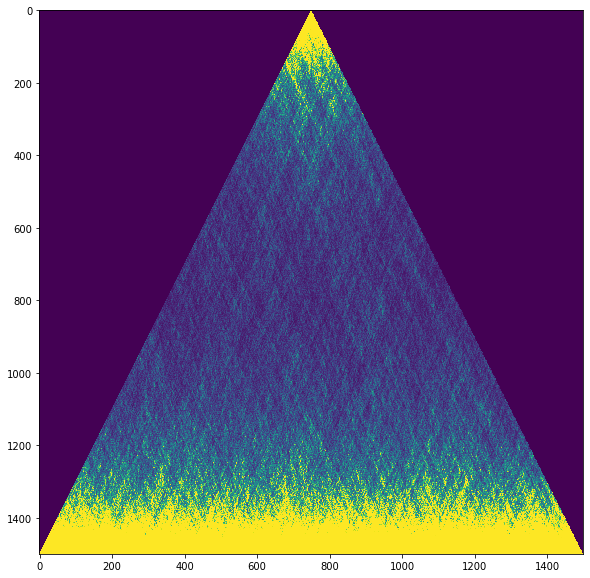}
    \caption{ Distribution of binary trees over an 1000-token sequence. Coloring shows the marginal probabilities of every span. Torch-Struct is an optimized collection of common CRF distributions used in NLP designed to integrate with deep learning frameworks. }
    \label{fig:my_label}
\end{figure}

\begin{table*}
\small 
\centering
\hspace*{-0.3cm}\begin{tabular}{llp{20mm}p{27mm}ccp{35mm}}
  \toprule
    Name & Structure ($\cal Z$) & Parts ($\cal P$) & Algorithm ($A(\ell)$) & LoC &  T/S & Sample Reference\\  

    \midrule
    Linear-Chain  & Labeled Chain & Edges $(TC^2)$ & Forward-Backward, \newline Viterbi & 20 & 390k &  \cite{lafferty2001conditional}  \\
    Factorial-HMM & Labeled Chains & Trans. $(LC^2)$ \newline Obs. $(TC^L)$ & Factorial F-B & 20 & 25k &  \cite{ghahramani1996factorial} \\ 
    Alignment & Alignment & Match $(NM)$ \newline Skips $(2 NM)$ & DTW, CTC, \newline Needleman-Wunsch & 50 & 13k & \cite{needleman1970general}\\ 
    Semi-Markov &  Seg. Labels & Edges $(NKC^2)$ & Segmental F-B & 30 & 87k & \cite{baum1966statistical} \newline \cite{sarawagi2005semi} \\ 
    Context-Free & Labeled Tree & CF Rules $(G)$ \newline Term. $(CN)$ & Inside-Outside CKY &   70 & 37k & \cite{kasami1966efficient}\\
    Simple CKY  & Labeled Tree &  Splits $(CN^2)$ & 0-th order CKY & 30 & 118k &\cite{kasami1966efficient} \\ \\
    Dependency & Proj. Tree & Arcs $(N^2)$ & Eisner Algorithm & 40  & 28k & \cite{eisner2000bilexical} \\ \\
    
    Dependency  (NP)&  Non-Proj. Tree & Arcs $(N^2)$ & Matrix-Tree \newline Chiu-Liu (MAP) & 40 &1.1m & \cite{koo2007structured} \newline \cite{mcdonald2005non} \\ 

    Auto-Regressive & Sequence & Prefix $(C^N)$ & Greedy Search, \newline Beam Search  & 60 & - & \cite{tillmann2003word} \\ 
    \bottomrule
    \end{tabular}
    \caption{\label{tab:models} Models and algorithms implemented in Torch-Struct. Notation is developed in Section~\ref{sec:tech}. \textit{Parts} are described in terms of sequence lengths $N,M$, label size $C$, segment length $K$, and layers / grammar size $L, G$. Lines of code (\textit{LoC}) is from the log-partition ($A(\ell)$) implementation. \textit{T/S} is the tokens per second of a batched computation, computed with batch 32, $N = 25, C = 20, K=5, L=3$  (K80 GPU run on Google Colab). }
\end{table*}

Structured prediction has played a key role in the history of natural language processing. Example methods include techniques for sequence labeling and segmentation \cite{lafferty2001conditional, sarawagi2005semi}, discriminative dependency and constituency parsing \cite{finkel2008efficient,mcdonald2005non}, unsupervised learning for labeling and alignment \cite{vogel1996hmm,goldwater2007fully}, approximate translation decoding with beam search \cite{tillmann2003word}, among  many others. 

In recent years, research into deep structured prediction has studied how these approaches can be integrated with neural networks and pretrained models. One line of work has utilized structured prediction as the final 
 final layer for deep models \cite{collobert2011natural,durrett2015neural}. Another has incorporated structured prediction within deep learning models, exploring  novel models for latent-structure learning, unsupervised learning, or model control~\cite{johnson2016composing,yogatama2016learning,wiseman2018learning}. We aspire to make both of these use-cases as easy to use as standard neural networks.  

The practical challenge of employing structured prediction is that many required algorithms are difficult to implement efficiently and correctly. Most projects reimplement custom versions of standard algorithms or focus particularly on a single well-defined model class. This research style makes it difficult to combine and try out new approaches, a problem that has compounded  with the complexity of research in deep structured prediction.

With this challenge in mind, we introduce Torch-Struct with three specific contributions: 
\begin{itemize}
    \item  \textit{Modularity}: models are represented as distributions with a standard flexible API integrated into 
    a deep learning framework. 
    
    \item \textit{Completeness}: a broad array of classical algorithms are implemented and new models can easily be added in Python.
    
    \item \textit{Efficiency}: implementations target computational/memory efficiency for GPUs and the backend includes extensions for optimization. 
\end{itemize}

\noindent In this system description, we first motivate the approach
taken by the library, then present a technical description of the methods used, and finally present several example use cases. 

\section{Related Work}

Several software libraries target structured prediction. Optimization tools, such as SVM-struct~\cite{joachims2008svmstruct}, focus on parameter estimation. Model libraries, such as CRFSuite~\cite{CRFsuite} or CRF++~\cite{kudo2005crf++},  implement inference for a fixed set of popular models, such as linear-chain CRFs. General-purpose inference libraries, such as PyStruct~\cite{muller2014pystruct} or TurboParser~\cite{martins2010turbo}, utilize external solvers for (primarily MAP) inference  such as integer linear programming solvers and ADMM. Probabilistic programming languages, for example languages that integrate with deep learning such as Pyro~\cite{bingham2019pyro}, allow for specification and inference over some discrete domains. Most ambitiously, inference libraries such as Dyna~\cite{eisner2004dyna} allow for declarative specifications of dynamic programming algorithms to support inference for generic algorithms. Torch-Struct takes a different approach and integrates a library of optimized structured distributions into a vectorized deep learning system. We begin by motivating this approach with a case study.


\section{Motivating Case Study}

While structured prediction is traditionally presented at the output layer, recent applications have deployed structured models broadly within neural networks \cite[inter alia]{johnson2016composing,DBLP:journals/corr/KimDHR17,yogatama2016learning}. Torch-Struct aims to encourage this general use case.

To illustrate, we consider a latent tree model. ListOps \cite{nangia2018listops} is a dataset of mathematical functions.  Each data point consists of a prefix expression $x$ and its result $y$, e.g.
\[ x= \mathrm{[\ MAX\ 2\ 9\ [\ MIN\ 4\ 7\ ]\ 0\  ]} \ \ y=9 \]
Models such as a flat RNN will fail to capture the hierarchical structure of this task. However, if a model can induce an explicit latent $z$, the parse tree of the expression, then the task is easy to learn by a tree-RNN model $p(y | x, z)$ \cite{yogatama2016learning,havrylov2019cooperative}. 

A popular approach is a latent-tree RL model which we briefly summarize. The objective is to 
maximize the probability of the correct prediction under the expectation of a prior tree model, $p(z|x ;\phi)$,
\[ O = \expect_{z \sim p(z | x;\phi)}[ \log p(y \ |\ z, x)]  \] 
 Computing the expectation is intractable so policy gradient is used. First a tree is sampled $\tilde{z} \sim p(z | x;\phi)$, then the gradient with respect to $\phi$ is approximated as,
\[ \frac{\partial }{\partial \phi} O  \approx (\log p(y \ | \tilde{z}, x )- b) (\frac{\partial }{\partial \phi} p(z | x; \phi)) \] 
\noindent where $b$ is a variance reduction baseline. 
A common choice is the self-critical baseline \cite{rennie2017self},  
\[   b = \log p(y \ |\ z^*, x) \mathrm{\ with\ } z^* = \arg\max_{z}p(z | x; \phi)\] 
Finally an entropy regularization term is added to the objective encourage exploration of different trees,
$ O + \lambda \entropy (p(z\ |\ x;\phi))$.

  


Even in this brief overview, we can see how complex a latent structured learning problem can be. To compute these terms, we need 5 different properties of the tree model $p(z\ | x; \phi)$:

\setlist[description]{font=\normalfont\itshape}

\begin{description}[itemsep=-2pt]
  \item[Sampling] Policy gradient, $\tilde{z} \sim p(z \ |\ x ; \phi)$
  \item[Density] Score policy samples, $p(z \ | \ x; \phi)$
  \item[Gradient] Backpropagation, $\frac{\partial }{\partial \phi} p(z\ |\ x; \phi)$
  \item[Argmax] Self-critical, $\arg\max_z p(z \ |\ x;\phi )$
  \item[Entropy] Objective regularizer, $\entropy(p(z\ |\ x;\phi))$
\end{description}
\noindent For structured models, each of these terms is non-trivial
to compute. A goal of Torch-Struct is to make it seamless to deploy structured models for these complex settings. To demonstrate this, Torch-Struct includes an implementation of this latent-tree approach. With a minimal amount of user code, the implementation achieves near perfect accuracy on the ListOps dataset. 



\section{Library Design}

The library design of Torch-Struct follows the distributions API used by both TensorFlow and PyTorch~\cite{dillon2017tensorflow}. 
For each structured model in the library, we define a conditional random field (CRF) distribution object. From a user's standpoint, this object provides all necessary distributional properties. Given log-potentials (scores) output from a deep network $\ell$, the user can request samples $z \sim \CRF(\ell)$, probabilities $\CRF(z;\ell)$, modes $\arg\max_z \CRF(\ell)$, or other distributional properties such as $\entropy(\CRF(\ell))$. The library is agnostic to how these are utilized, and when possible, they allow for backpropagation  to update the input network. The same distributional object can be used for standard output prediction as for more complex operations like attention or reinforcement learning.


\begin{figure}
    \centering
    \includegraphics[trim=0.7cm 0.8cm 0 0, clip, clip,width=0.5\linewidth]{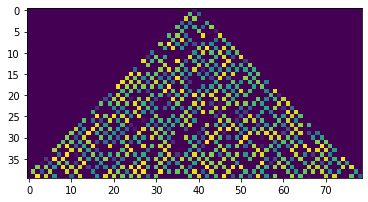}\includegraphics[trim=0.7cm 0.8cm 0 0, clip, width=0.5\linewidth]{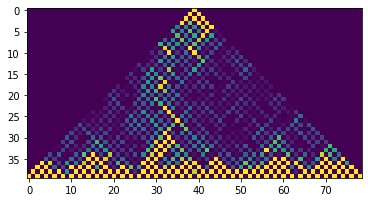}
    \includegraphics[trim=0.7cm 0.8cm 0 0, clip, width=0.5\linewidth]{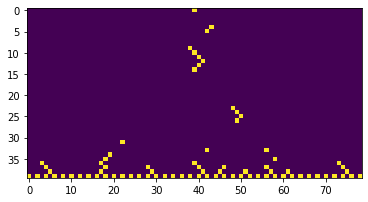}\includegraphics[trim=0.7cm 0.8cm 0 0, clip,width=0.5\linewidth]{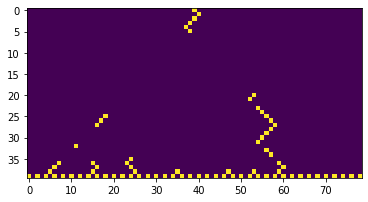}
    \caption{Latent Tree CRF example. (a) Log-potentials $\ell$ for each part/span. (b) Marginals for   $\CRF(\ell)$ computed by backpropagation. (c) Mode tree $\arg\max_{z} \CRF(z; \ell)$. (d) Sampled tree $z \sim \CRF(\ell)$. }
    \label{fig:deptrees}
\end{figure}

Figure~\ref{fig:deptrees} demonstrates this API for a binary tree CRF over an ordered sequence, such as $p(z \ | \ y ;\phi)$ from the previous section. The distribution takes in log-potentials $\ell$ which score each possible span in the input. The distribution converts these to probabilities of a specific tree. This distribution can be queried for predicting over the set of trees, sampling a tree for model structure, or even computing entropy over all trees. 

Table~\ref{tab:models} shows all of the structures and distributions implemented in Torch-Struct. While each is internally implemented using different specialized algorithms and optimizations, from the user's perspective they all utilize the same external distributional API, and pass a generic set of distributional tests.\footnote{The test suite for each distribution enumerates over all structures to ensure that properties hold. While this is intractable for large spaces, it can be done for small sets and was extremely useful for development.} This approach hides the internal complexity of the inference procedure, while giving the user full access to the model. 

\section{Technical Approach}
\label{sec:tech}
\subsection{Conditional Random Fields}

We now describe the technical approach underlying the library. 
To establish notation first consider the implementation of a categorical distribution, \Cat($\ell$), with one-hot categories $z$ with $z_i = 1$ from a set $\cal Z$ and probabilities given by the softmax,
\[\Cat({z} ; \mathbf{\ell}) = \frac{\exp ({z} \cdot \ell)}{\sum_{{z}' \in \cal Z} \exp ({z'} \cdot \ell) } = \frac{\exp \ell_i}{ \sum_{j=1}^K \exp \ell_j}\]
\noindent Define the log-partition as $A(\ell) = \LSE(\ell)$, i.e.  log of the denominator, where $\LSE$ is the log-sum-exp operator. Computing probabilities or sampling from this distribution, requires enumerating $\cal Z$ to compute the log-partition $A$. A useful identity is that derivatives of $A$ yield category probabilities, 
$$p(z_i = 1) = \frac{\exp \ell_i}{\sum_{j=1}^n \exp \ell_j } = \frac{\partial}{\partial \ell_i} A(\ell) $$ 
Other distributional properties can be similarly extracted from variants of the log-partition. For instance, define $A^*(\ell) = \log \max_{j=1}^K \exp \ell_j$ then\footnote{This is a subgradient identity, but we observe that libraries like PyTorch will default to this value.}:
$\mathbb{I}(z^*_i = 1) =  \frac{\partial}{\partial \ell_i} A^*(\ell) $.







Conditional random fields, \textsc{CRF}($\ell$), extend the softmax to combinatorial spaces where ${\cal Z}$ is exponentially sized.  Each $z$, is now represented as a binary vector over polynomial-sized set of \textit{parts}, $\cal P$, i.e. ${\cal Z} \subset \{0, 1\}^{|\cal P|}$. Similarly log-potentials are now defined over parts $\ell \in \reals^{|\cal P|}$. For instance, in Figure~\ref{fig:deptrees} each span is a part and the $\ell$ vector is shown in the top-left figure.
Define the probability of a structure $z$ as,
\[ \mathrm{CRF}(z ;\ell) = \frac{\exp z \cdot \ell}{\sum_{z'} \exp z'  \cdot \ell} =\frac{\exp \sum_{p} \ell_p z_p}{\sum_{z'} \exp \sum_{p }  \ell_p z'_p} \]
Computing probabilities or sampling from this distribution, requires computing the log-partition term $A$. 
In general computing this term is now intractable, however 
for many core algorithms in NLP there are exist efficient combinatorial algorithms for this term (as enumerated in Table~\ref{tab:models}). 

Derivatives of the log-partition again provide distributional properties. 
For instance, the marginal probabilities of parts are given by,  
$$p(z_p = 1) =  \frac{\exp \sum_{z:z_p = 1} z \cdot \ell}{\sum_{z' \in } \exp z'  \cdot \ell} = \frac{\partial}{\partial \ell_p} A(\ell)  $$ 
Similarly derivatives of $A^*$ correspond to whether a part appears in the argmax structure. 
$\mathbb{I}(z^*_p = 1)  = \frac{\partial}{\partial \ell_p} A^*(\ell) $.

While these gradient identities are well-known \cite{eisner2016inside}, they are not commonly deployed. Computing CRF properties is typically done through two-step specialized algorithms, such as forward-backward, inside-outside, or similar variants such as viterbi-backpointers \cite{jurafsky2014speech}.  In our experiments, we found that using these identities with auto-differentiation on GPU was often faster, and much simpler, than custom two-pass approaches.
Torch-Struct is thus designed around using gradients for distributional computations. 

\subsection{Dynamic Programming and Semirings}

\begin{table}
\small 
\begin{tabular}{lcccc}
  \toprule
    Name & Ops ($\bigoplus, \otimes$) &  Backprop & Gradients \\  
    \midrule
    Log & $\LSE, +$ &  $\Delta $ & $p(z_p = 1)$\\
    Max & $\max, +$ &  $\Delta $& $\arg\max_z $\\
    K-Max & $k\max, +$ & $\Delta $&  K-Argmax\\ 
    Sample & $\LSE, +$ &  $\sim$ &  $z\sim \CRF(\ell)$\\
    K-Sample & $\LSE, +$ &  $\sim$ &  K-Samples\\
    \midrule
    Count& $\sum, \times$ &  & \\
    Entropy ( $\entropy $) &  \multicolumn{3}{c}{See \cite{li2009first}} & \\
    Exp. &  \multicolumn{3}{c}{See \cite{li2009first}} \\
    Sparsemax &  \multicolumn{3}{c}{See \cite{mensch2018differentiable}}   \\
    \bottomrule
    \end{tabular}
    
    \includegraphics[clip, trim=2cm 0 0 2cm , width=\linewidth]{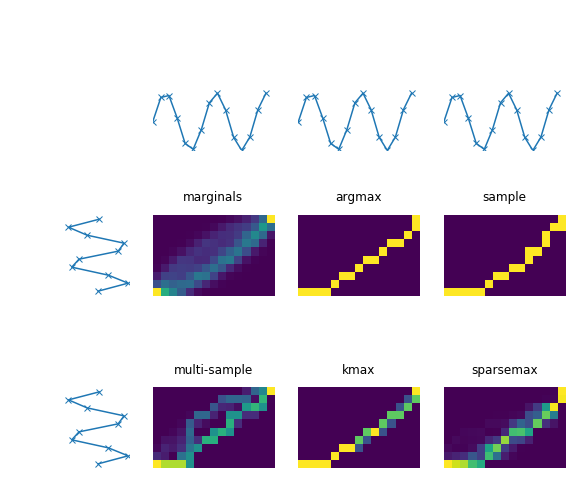}
\vspace{-0.7cm}

    \caption{\label{tab:semi} (Top) Semirings implemented in Torch-Struct. 
    \textit{Backprop/Gradients} gives overridden backpropagation computation and value computed by this combination.
    (Bot) Example of gradients from different semirings on sequence alignment with dynamic time warping.}
\end{table}

Torch-Struct is a collection of generic algorithms for CRF inference. Each CRF distribution object, $\CRF(\ell)$, is constructed by providing 
$\ell \in \reals^{|{\cal P}|}$ where the parts $\cal P$ are specific to the type of distribution. Internally, each 
distribution is implemented through a single Python function for computing the log-partition function $A(\ell)$. From this function, the library uses auto-differentiation and the identities from the previous section, to define a complete distribution object. 
The core models implemented by the library are shown in Table~\ref{tab:models}. 




 \begin{figure*}
\includegraphics[width=0.33\textwidth]{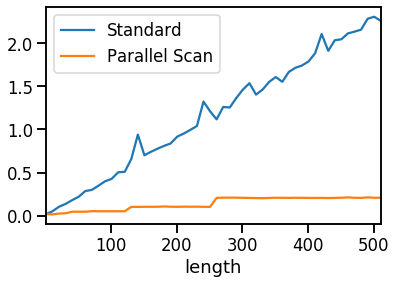}
\includegraphics[width=0.31\textwidth]{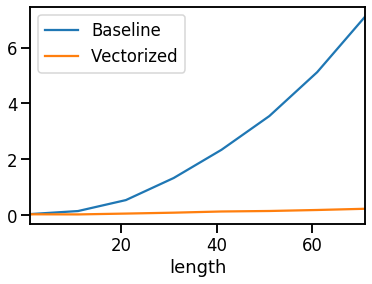}
\includegraphics[width=0.325\textwidth]{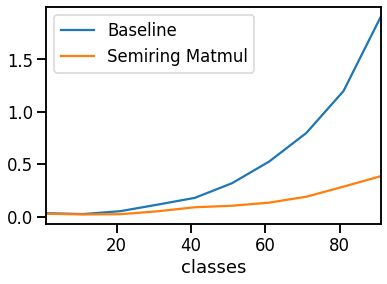}
\caption{Speed impact of optimizations. Time is given in seconds for 10 runs with batch 16 executed on Google Colab. (a) Speed of a linear-chain forward with 20 classes for lengths up to 500. Compares left-to-right ordering to parallel scan. (b) Speed of CKY inside with lengths up to 80. Compares inner loop versus vectorization. (c) Speed of linear-chain forward of length 20 with up to 100 classes. Compares broadcast-reduction versus CUDA semiring kernel.
(Baseline memory is exhausted after 100 classes.)
}
\label{fig:speed}
\end{figure*}

To make the approach concrete, we consider the example of a  linear-chain CRF. 

\begin{center}    
\begin{tikzpicture}
\node[latent](a){$z_1$};
\node[latent, right = of a](b){$z_2$};
\node[latent, right = of b](c){$z_3$};
\draw (a) -- (b) -- (c);
\end{tikzpicture}
\end{center}

The model has $C$ labels per node with a length $T=2$ edges utilizing a first-order linear-chain (Markov) model. This model has $2\times C \times C$ parts corresponding to edges in the chain, and thus requires $\ell \in \reals^{2\times C \times C}$. The log-partition function $A(\ell)$ factors into two reduce computations,
\begin{eqnarray*}
A(\ell) &=& \log  \sum_{c_{3}, c_{2}} \exp  \ell_{2,c_{2},c_{3}} \sum_{c_1} \exp \ell_{1,c_{1},c_2} \\ 
 &=&  \LSE_{c_3, c_2}[ \ell_{2,c_{2},c_{3}} + [\LSE_{c_1}  \ell_{1,c_{1},c_2}]] 
 \end{eqnarray*}
 Computing this function left-to-right using dynamic programming 
 yield the standard \textit{forward} algorithm for sequence models. As we have seen, the gradient with respect to $\ell$ produces marginals for each part, i.e. the probability of a specific labeled edge.

We can further extend the same function to support generic \textit{semiring} dynamic programming~\cite{wgoodman1999semiring}. A semiring is defined by a pair $(\oplus, \otimes)$ with commutative $\oplus$, distribution, and appropriate identities. The log-partition utilizes 
$\oplus, \otimes = \LSE, +$, but we can substitute alternatives.
\begin{eqnarray*}
A(\ell) &=&   \bigoplus_{c_3, c_2}[ \ell_{2,c_{2},c_{3}} \otimes [\bigoplus_{c_1}  \ell_{1,c_{1},c_2}]] 
 \end{eqnarray*}
  For instance, utilizing the log-max semiring $(\max, +)$ in the forward algorithm yields the max score. As we have seen, its gradient with respect to $\ell$ is the argmax sequence, negating the need for a separate argmax (Viterbi) algorithm. 
Some distributional properties cannot be computed directly through gradient identities but still use a forward-backward style compute structure. For instance, sampling requires first computing the log-partition term and then sampling each part, (forward filtering / backward sampling). We can compute this value by overriding each backpropagation operation for the $\bigoplus$ to instead compute a sample.

Table~\ref{tab:semi} shows the set of semirings and backpropagation steps for computing different terms of interest. We note that many of the terms necessary in the case-study can be computed with variant semirings, negating the need for specialized algorithms.

\section{Optimizations}

Torch-Struct aims for computational and memory efficiency. 
Implemented naively, dynamic programming algorithms in Python are 
prohibitively slow. As such Torch-Struct provides key primitives
to help batch and vectorize these algorithms to take advantage
of GPU computation and to minimize the overhead of backpropagating 
through chart-based dynamic programmming. Figure~\ref{fig:speed} shows the impact of these optimizations on the core algorithms.

\paragraph{a) Parallel Scan Inference}

\begin{figure}
    \centering
    \begin{tikzpicture}[level distance=1cm]
    \Tree [ .$A(\ell)$ [ .$\bigoplus \otimes$ [.$\bigoplus \otimes$ $\ell_{1, \cdot, \cdot}$ $\ell_{2, \cdot, \cdot}$ ] [ .$\bigoplus \otimes$ $\ell_{3, \cdot, \cdot}$ $\ell_{4, \cdot, \cdot}$ ] ] [.$\bigoplus \otimes$  [ .$\bigoplus \otimes$ $\ell_{5, \cdot, \cdot}$ $\ell_{6,\cdot, \cdot}$ ] [ .$\bigoplus \otimes$ $\ell_{7,\cdot, \cdot}$ $I$ ] ] ]
    \end{tikzpicture}
    \caption{Parallel scan implementation of the linear-chain CRF inference algorithm. Here $\bigoplus \otimes$ represents a semiring matrix operation and $I$ is padding. }
    \label{fig:pscan}
\end{figure}
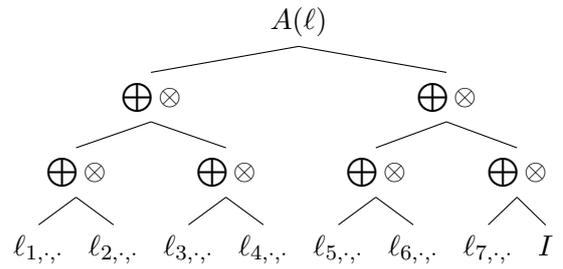

The commutative properties of semiring algorithms allow flexibility in the order in which we compute $A(\ell)$.
Typical implementations of dynamic programming algorithms are serial in the length of the sequence. On parallel hardware, an appealing approach is a parallel scan ordering~\cite{sarkka2019temporal}, typically used for computing prefix sums. To compute, $A(\ell)$ in this manner we first pad the sequence length $T$ out to the nearest power of two, and then compute a balanced parallel tree over the parts, shown in Figure~\ref{fig:pscan}. Concretely each node layer would compute a semiring matrix multiplication, e.g. $ \bigoplus_c \ell_{t, \cdot , c} \otimes \ell_{t', c, \cdot}$.
Under this approach, we only need $O(\log N)$ steps in Python and can use parallel GPU operations for the rest. Similar parallel approach can also be used for computing sequence alignment and semi-Markov models. 

\paragraph{b) Vectorized Parsing}

Computational complexity is even more of an issue for parsing algorithms, which cannot be as easily parallelized. The log-partition for parsing is computed with the Inside algorithm. This algorithm 
must compute each width from 1 through T in serial; however it is important to parallelize each inner step. 
Assuming we have computed all inside spans of width less than $d$, computing the inside span of width $d$ requires computing for all $i$,
\begin{eqnarray*}
C[i, i+d] &=& \bigoplus_{j=i}^{i+d-1} C[i, j] \otimes C[j+1, i+d] 
\end{eqnarray*}
 In order to vectorize this loop over $i, j$, we reindex the chart. Instead of using a single chart $C$, we split it into two parts: one right-facing $C_r[i, d] = C[i, i+d]$ and one left facing, $C_l[i+d, T-d] = C[i, i+d]$. After this reindexing, the update can be written.
$$C_r[i, d] = \bigoplus_{j=1}^{j-1} C_r[i, j] \otimes C_l[i+d, T- d+j] $$
Unlike the original, this formula can easily be computed as a vectorized semiring dot product. This allows use to compute 
 $C_r[\cdot, d]$ in one operation. Variants of this same approach can be used for all the parsing models employed.

\paragraph{c) Semiring Matrix Operations}
The two previous optimizations reduce most of the cost to semiring matrix multiplication. In the specific case of the $(\sum, \times)$ semiring these can be computed very efficiently using matrix multiplication, which is highly-tuned on GPU hardware. Unfortunately for other semirings, such as log and max, these operations are either slow or very memory inefficient. For instance, for matrices $T$ and $U$ of sized $N \times M$ and $M \times O$, we can broadcast with $\otimes$ to a tensor of size $N \times M \times O$ and then reduce dim $M$ by $\bigoplus$ at a huge memory cost. To avoid this issue, we implement custom CUDA kernels targeting fast and memory efficient tensor operations. For log, this corresponds to computing, $$V_{m, o} = \log \sum_{n} \exp(T_{m,n} + U_{n, o} - q) + q$$ 
where $q = \max_n T_{m,n} + U_{n, o}$. To optimize this operation on GPU we utilize the TVM language \cite{chen2018tvm} to layout the CUDA loops and tune it to hardware.  

\section{Conclusion and Future Work}

We present Torch-Struct, a library for deep structured prediction. The library 
achieves modularity through its adoption of a generic distributional API, 
completeness by utilizing CRFs and semirings to make it easy to add new algorithms, 
and efficiency through core optimizations to vectorize important dynamic programming 
steps.  In addition to the problems discussed so far, Torch-Struct also includes several other example implementations including supervised dependency parsing with BERT, 
unsupervised tagging, structured attention, and connectionist temporal classification (CTC) for speech. The full library is available at \url{https://github.com/harvardnlp/pytorch-struct}.

In the future, we hope to support research and production 
applications employing structured models. We also believe the library provides a strong  foundation for building generic tools for interpretablity, control, and visualization through its probabilistic API. Finally, we hope to explore further optimizations to make core algorithms competitive with  highly-optimized neural network components. 

\section*{Acknowledgements}

We thank Yoon Kim, Xiang Lisa Li, Sebastian Gehrmann, Yuntian Deng, and Justin Chiu for discussion and feedback on the project. The project was supported by NSF CAREER 1845664, NSF 1901030, and research awards by Sony and AWS.

\bibliographystyle{acl_natbib}
\bibliography{acl2020,paperpile,refs}

\begin{thebibliography}{37}
\expandafter\ifx\csname natexlab\endcsname\relax\def\natexlab#1{#1}\fi

\bibitem[{Baum and Petrie(1966)}]{baum1966statistical}
Leonard~E Baum and Ted Petrie. 1966.
\newblock Statistical inference for probabilistic functions of finite state
  markov chains.
\newblock \emph{The annals of mathematical statistics}, 37(6):1554--1563.

\bibitem[{Bingham et~al.(2019)Bingham, Chen, Jankowiak, Obermeyer, Pradhan,
  Karaletsos, Singh, Szerlip, Horsfall, and Goodman}]{bingham2019pyro}
Eli Bingham, Jonathan~P Chen, Martin Jankowiak, Fritz Obermeyer, Neeraj
  Pradhan, Theofanis Karaletsos, Rohit Singh, Paul Szerlip, Paul Horsfall, and
  Noah~D Goodman. 2019.
\newblock Pyro: Deep universal probabilistic programming.
\newblock \emph{The Journal of Machine Learning Research}, 20(1):973--978.

\bibitem[{Chen et~al.(2018)Chen, Moreau, Jiang, Shen, Yan, Wang, Hu, Ceze,
  Guestrin, and Krishnamurthy}]{chen2018tvm}
Tianqi Chen, Thierry Moreau, Ziheng Jiang, Haichen Shen, Eddie Yan, Leyuan
  Wang, Yuwei Hu, Luis Ceze, Carlos Guestrin, and Arvind Krishnamurthy. 2018.
\newblock Tvm: end-to-end optimization stack for deep learning.
\newblock \emph{arXiv preprint arXiv:1802.04799}.

\bibitem[{Collobert et~al.(2011)Collobert, Weston, Bottou, Karlen, Kavukcuoglu,
  and Kuksa}]{collobert2011natural}
Ronan Collobert, Jason Weston, L{\'e}on Bottou, Michael Karlen, Koray
  Kavukcuoglu, and Pavel Kuksa. 2011.
\newblock Natural language processing (almost) from scratch.
\newblock \emph{Journal of machine learning research}, 12(Aug):2493--2537.

\bibitem[{Dillon et~al.(2017)Dillon, Langmore, Tran, Brevdo, Vasudevan, Moore,
  Patton, Alemi, Hoffman, and Saurous}]{dillon2017tensorflow}
Joshua~V Dillon, Ian Langmore, Dustin Tran, Eugene Brevdo, Srinivas Vasudevan,
  Dave Moore, Brian Patton, Alex Alemi, Matt Hoffman, and Rif~A Saurous. 2017.
\newblock Tensorflow distributions.
\newblock \emph{arXiv preprint arXiv:1711.10604}.

\bibitem[{Durrett and Klein(2015)}]{durrett2015neural}
Greg Durrett and Dan Klein. 2015.
\newblock Neural crf parsing.
\newblock \emph{arXiv preprint arXiv:1507.03641}.

\bibitem[{Eisner(2000)}]{eisner2000bilexical}
Jason Eisner. 2000.
\newblock Bilexical grammars and their cubic-time parsing algorithms.
\newblock In \emph{Advances in probabilistic and other parsing technologies},
  pages 29--61. Springer.

\bibitem[{Eisner(2016)}]{eisner2016inside}
Jason Eisner. 2016.
\newblock Inside-outside and forward-backward algorithms are just backprop
  (tutorial paper).
\newblock In \emph{Proceedings of the Workshop on Structured Prediction for
  NLP}, pages 1--17.

\bibitem[{Eisner et~al.(2004)Eisner, Goldlust, and Smith}]{eisner2004dyna}
Jason Eisner, Eric Goldlust, and Noah~A Smith. 2004.
\newblock Dyna: A declarative language for implementing dynamic programs.
\newblock In \emph{Proceedings of the ACL 2004 on Interactive poster and
  demonstration sessions}, pages 32--es.

\bibitem[{Finkel et~al.(2008)Finkel, Kleeman, and
  Manning}]{finkel2008efficient}
Jenny~Rose Finkel, Alex Kleeman, and Christopher~D Manning. 2008.
\newblock Efficient, feature-based, conditional random field parsing.
\newblock In \emph{Proceedings of ACL-08: HLT}, pages 959--967.

\bibitem[{Ghahramani and Jordan(1996)}]{ghahramani1996factorial}
Zoubin Ghahramani and Michael~I Jordan. 1996.
\newblock Factorial hidden markov models.
\newblock In \emph{Advances in Neural Information Processing Systems}, pages
  472--478.

\bibitem[{Goldwater and Griffiths(2007)}]{goldwater2007fully}
Sharon Goldwater and Tom Griffiths. 2007.
\newblock A fully bayesian approach to unsupervised part-of-speech tagging.
\newblock In \emph{Proceedings of the 45th annual meeting of the association of
  computational linguistics}, pages 744--751.

\bibitem[{Goodman(1999)}]{wgoodman1999semiring}
Joshua Goodman. 1999.
\newblock Semiring parsing.
\newblock \emph{Computational Linguistics}, 25(4):573--605.

\bibitem[{Havrylov et~al.(2019)Havrylov, Kruszewski, and
  Joulin}]{havrylov2019cooperative}
Serhii Havrylov, Germ{\'a}n Kruszewski, and Armand Joulin. 2019.
\newblock Cooperative learning of disjoint syntax and semantics.
\newblock \emph{arXiv preprint arXiv:1902.09393}.

\bibitem[{Joachims(2008)}]{joachims2008svmstruct}
Thorsten Joachims. 2008.
\newblock Svmstruct: Support vector machine for complex outputs.

\bibitem[{Johnson et~al.(2016)Johnson, Duvenaud, Wiltschko, Adams, and
  Datta}]{johnson2016composing}
Matthew~J Johnson, David~K Duvenaud, Alex Wiltschko, Ryan~P Adams, and
  Sandeep~R Datta. 2016.
\newblock Composing graphical models with neural networks for structured
  representations and fast inference.
\newblock In \emph{Advances in neural information processing systems}, pages
  2946--2954.

\bibitem[{Jurafsky and Martin(2014)}]{jurafsky2014speech}
Dan Jurafsky and James~H Martin. 2014.
\newblock Speech and language processing. vol. 3.

\bibitem[{Kasami(1966)}]{kasami1966efficient}
Tadao Kasami. 1966.
\newblock An efficient recognition and syntax-analysis algorithm for
  context-free languages.
\newblock \emph{Coordinated Science Laboratory Report no. R-257}.

\bibitem[{Kim et~al.(2017)Kim, Denton, Hoang, and
  Rush}]{DBLP:journals/corr/KimDHR17}
Yoon Kim, Carl Denton, Luong Hoang, and Alexander~M. Rush. 2017.
\newblock \href {http://arxiv.org/abs/1702.00887} {Structured attention
  networks}.
\newblock \emph{CoRR}, abs/1702.00887.

\bibitem[{Koo et~al.(2007)Koo, Globerson, Carreras~P{\'e}rez, and
  Collins}]{koo2007structured}
Terry Koo, Amir Globerson, Xavier Carreras~P{\'e}rez, and Michael Collins.
  2007.
\newblock Structured prediction models via the matrix-tree theorem.
\newblock In \emph{Joint Conference on Empirical Methods in Natural Language
  Processing and Computational Natural Language Learning (EMNLP-CoNLL)}, pages
  141--150.

\bibitem[{Kudo(2005)}]{kudo2005crf++}
Taku Kudo. 2005.
\newblock Crf++: Yet another crf toolkit.
\newblock \emph{http://crfpp. sourceforge. net/}.

\bibitem[{Lafferty et~al.(2001)Lafferty, McCallum, and
  Pereira}]{lafferty2001conditional}
John Lafferty, Andrew McCallum, and Fernando~CN Pereira. 2001.
\newblock Conditional random fields: Probabilistic models for segmenting and
  labeling sequence data.

\bibitem[{Li and Eisner(2009)}]{li2009first}
Zhifei Li and Jason Eisner. 2009.
\newblock First-and second-order expectation semirings with applications to
  minimum-risk training on translation forests.
\newblock In \emph{Proceedings of the 2009 Conference on Empirical Methods in
  Natural Language Processing: Volume 1-Volume 1}, pages 40--51. Association
  for Computational Linguistics.

\bibitem[{Martins et~al.(2010)Martins, Smith, Xing, Aguiar, and
  Figueiredo}]{martins2010turbo}
Andr{\'e}~FT Martins, Noah~A Smith, Eric~P Xing, Pedro~MQ Aguiar, and
  M{\'a}rio~AT Figueiredo. 2010.
\newblock Turbo parsers: Dependency parsing by approximate variational
  inference.
\newblock In \emph{Proceedings of the 2010 Conference on Empirical Methods in
  Natural Language Processing}, pages 34--44. Association for Computational
  Linguistics.

\bibitem[{McDonald et~al.(2005)McDonald, Pereira, Ribarov, and
  Haji{\v{c}}}]{mcdonald2005non}
Ryan McDonald, Fernando Pereira, Kiril Ribarov, and Jan Haji{\v{c}}. 2005.
\newblock Non-projective dependency parsing using spanning tree algorithms.
\newblock In \emph{Proceedings of the conference on Human Language Technology
  and Empirical Methods in Natural Language Processing}, pages 523--530.
  Association for Computational Linguistics.

\bibitem[{Mensch and Blondel(2018)}]{mensch2018differentiable}
Arthur Mensch and Mathieu Blondel. 2018.
\newblock Differentiable dynamic programming for structured prediction and
  attention.
\newblock \emph{arXiv preprint arXiv:1802.03676}.

\bibitem[{M{\"u}ller and Behnke(2014)}]{muller2014pystruct}
Andreas~C M{\"u}ller and Sven Behnke. 2014.
\newblock Pystruct: learning structured prediction in python.
\newblock \emph{The Journal of Machine Learning Research}, 15(1):2055--2060.

\bibitem[{Nangia and Bowman(2018)}]{nangia2018listops}
Nikita Nangia and Samuel~R Bowman. 2018.
\newblock Listops: A diagnostic dataset for latent tree learning.
\newblock \emph{arXiv preprint arXiv:1804.06028}.

\bibitem[{Needleman and Wunsch(1970)}]{needleman1970general}
Saul~B Needleman and Christian~D Wunsch. 1970.
\newblock A general method applicable to the search for similarities in the
  amino acid sequence of two proteins.
\newblock \emph{Journal of molecular biology}, 48(3):443--453.

\bibitem[{Okazaki(2007)}]{CRFsuite}
Naoaki Okazaki. 2007.
\newblock \href {http://www.chokkan.org/software/crfsuite/} {Crfsuite: a fast
  implementation of conditional random fields (crfs)}.

\bibitem[{Rennie et~al.(2017)Rennie, Marcheret, Mroueh, Ross, and
  Goel}]{rennie2017self}
Steven~J Rennie, Etienne Marcheret, Youssef Mroueh, Jerret Ross, and Vaibhava
  Goel. 2017.
\newblock Self-critical sequence training for image captioning.
\newblock In \emph{Proceedings of the IEEE Conference on Computer Vision and
  Pattern Recognition}, pages 7008--7024.

\bibitem[{Sarawagi and Cohen(2005)}]{sarawagi2005semi}
Sunita Sarawagi and William~W Cohen. 2005.
\newblock Semi-markov conditional random fields for information extraction.
\newblock In \emph{Advances in neural information processing systems}, pages
  1185--1192.

\bibitem[{S{\"a}rkk{\"a} and
  Garc{\'\i}a-Fern{\'a}ndez(2019)}]{sarkka2019temporal}
Simo S{\"a}rkk{\"a} and {\'A}ngel~F Garc{\'\i}a-Fern{\'a}ndez. 2019.
\newblock Temporal parallelization of bayesian filters and smoothers.
\newblock \emph{arXiv preprint arXiv:1905.13002}.

\bibitem[{Tillmann and Ney(2003)}]{tillmann2003word}
Christoph Tillmann and Hermann Ney. 2003.
\newblock Word reordering and a dynamic programming beam search algorithm for
  statistical machine translation.
\newblock \emph{Computational linguistics}, 29(1):97--133.

\bibitem[{Vogel et~al.(1996)Vogel, Ney, and Tillmann}]{vogel1996hmm}
Stephan Vogel, Hermann Ney, and Christoph Tillmann. 1996.
\newblock Hmm-based word alignment in statistical translation.
\newblock In \emph{Proceedings of the 16th conference on Computational
  linguistics-Volume 2}, pages 836--841. Association for Computational
  Linguistics.

\bibitem[{Wiseman et~al.(2018)Wiseman, Shieber, and Rush}]{wiseman2018learning}
Sam Wiseman, Stuart~M Shieber, and Alexander~M Rush. 2018.
\newblock Learning neural templates for text generation.
\newblock \emph{arXiv preprint arXiv:1808.10122}.

\bibitem[{Yogatama et~al.(2016)Yogatama, Blunsom, Dyer, Grefenstette, and
  Ling}]{yogatama2016learning}
Dani Yogatama, Phil Blunsom, Chris Dyer, Edward Grefenstette, and Wang Ling.
  2016.
\newblock Learning to compose words into sentences with reinforcement learning.
\newblock \emph{arXiv preprint arXiv:1611.09100}.

\end{thebibliography}
\end{document}